\documentclass[pdflatex, ,sn-mathphys-num]{sn-jnl}% Math and Physical Sciences Numbered Reference Style
\usepackage{graphicx}%
\usepackage{multirow}%
\usepackage{amsmath,amssymb,amsfonts}%
\usepackage{amsthm}%
\usepackage{mathrsfs}%
\usepackage[title]{appendix}%
\usepackage{xcolor}%
\usepackage{textcomp}%
\usepackage{manyfoot}%
\usepackage{booktabs}%
\usepackage{algorithm}%
\usepackage{algorithmicx}%
\usepackage{algpseudocode}%
\usepackage{listings}%
\usepackage{float}
\usepackage{tabularx}
\usepackage{adjustbox}
\usepackage{comment}

\definecolor{red1}{RGB}{218, 65, 61}
\definecolor{gray1}{RGB}{156,169,181}
\definecolor{blue1}{RGB}{68,102,165}

\algnewcommand\algorithmicforeach{\textbf{for each}}
\algdef{S}[FOR]{ForEach}[1]{\algorithmicforeach\ #1\ \algorithmicdo}
\begin{document}

\title[Vectorized FlashAttention with Low-cost Exponential Computation in RISC-V Vector Processors]{Vectorized FlashAttention with Low-cost Exponential Computation in RISC-V Vector Processors}

\author*{\fnm{Vasileios} \sur{Titopoulos}}\email{vtitopou@ee.duth.gr}

\author{\fnm{Kosmas} \sur{Alexandridis}}\email{koalexan@ee.duth,gr}

\author{\fnm{Giorgos} \sur{Dimitrakopoulos}}\email{dimitrak@ee.duth.gr}

\affil{\orgdiv{Electrical and Computer Engineering}, \orgname{Democritus University of Thrace}, \orgaddress{\street{Kimmeria Campus}, \city{Xanthi}, \postcode{67100}, \country{Greece}}}

\abstract{Attention is a core operation in numerous machine learning and artificial intelligence models. This work focuses on the acceleration of attention kernel using FlashAttention algorithm, in vector processors, particularly those based on the RISC-V instruction set architecture (ISA). This work represents the first effort to vectorize FlashAttention, minimizing scalar code and simplifying the computational complexity of evaluating exponentials needed by softmax used in attention. By utilizing a low-cost approximation for exponentials in floating-point arithmetic, we reduce the cost of computing the exponential function without the need to extend baseline vector ISA with new custom instructions. Also, appropriate tiling strategies are explored with the goal to improve memory locality. Experimental results highlight the scalability of our approach, demonstrating significant performance gains with the vectorized implementations when processing attention layers in practical applications.}

\keywords{Transformers, Attention, Vector processors, RISC-V}

\maketitle

\section{Introduction}\label{sec1}
The transformer architecture has brought major progress to machine learning, especially in large language models (LLMs). It has been very effective for sequence-based tasks like natural language processing~\cite{t5} and time-series analysis~\cite{tran_surv}. One of the key reasons for this success is the attention mechanism, which allows models to focus on the most relevant parts of the input~\cite{base_attn}. Thanks to this, transformers can capture long-distance relationships in data, where recurrent neural networks have often struggled with.

Attention works by giving different levels of importance to each part of the input, depending on how useful it is at each step. Instead of treating all parts equally, the model scores each part based on relevance. These scores help the model give more weight to important information and less to irrelevant data, no matter where it appears in the sequence.

However, this ability comes with a high computational cost. The standard attention mechanism has quadratic complexity, which means its time and memory usage increase rapidly with longer sequences~\cite{longformer}. This creates a performance bottleneck and limits the use of transformers on long inputs. As a result, tasks like document summarization and code generation, which require understanding long contexts, can be difficult to handle efficiently.

To address this challenge, researchers have looked into different ways to reduce the high computational cost of attention. One common strategy is to simplify the attention calculation by using sparse~\cite{sparse_attn}, linear~\cite{lin_attn}, or low-rank~\cite{low_rank_attn_2020} attention methods. These techniques aim to improve efficiency while keeping accuracy high. In addition, specialized hardware accelerators~\cite{a3,lazy_softmax, isvlsi} have been created to speed up key parts of the attention process, such as matrix operations~\cite{cosa} and softmax calculations~\cite{softermax}, which helps improve speed and scalability.

FlashAttention~\cite{fa,fa2,nsquared}, originally designed for GPUs, efficiently addresses computational bottlenecks in standard attention mechanisms. By performing attention computations in tiles rather than storing the entire attention matrix, it reduces memory requirements thus enhancing both memory efficiency and computational performance.

In this work, we combine the high performance of the FlashAttention algorithm with the cost and energy efficiency of vector processors to address the growing computational demands of attention in large models. While FlashAttention significantly improves computation of attention on GPUs, its benefits have not yet been extended to vector architectures, which are well-suited for parallel data processing with lower hardware complexity and power consumption. To bridge this gap, we propose the first design of a vectorized FlashAttention algorithm, targeting the RISC-V Vector ISA extension. Our goal is to enable efficient, scalable, and open-source attention computation on vector processors, making it more accessible for resource-constrained and specialized hardware platforms.

The contributions of this work can be summarized as follows:

\begin{itemize}
\item
We present for the first time, to the best of our knowledge, a fully vectorized implementation of the FlashAttention algorithm, composed entirely of vector instructions, without relying on scalar operations. This ensures maximum utilization of data-level parallelism provided by vector processors and allows efficient execution on architectures that support the RISC-V Vector ISA.

\item
To support the softmax operation, we integrate a simplified method for computing the exponential function using standard floating-point arithmetic. Our approach is tailored for data-parallel execution and FlashAttention algorithm, without requiring any custom hardware instructions or ISA extensions and without compromising accuracy of representative LLM applications.

\item
We design a tiling strategy that leverages the core idea of FlashAttention that processes attention in small, memory-efficient blocks to suit vectorized execution. Our tiling approach maximizes data reuse, minimizes memory access latency, and fits naturally within the vector register file constraints of typical RISC-V vector processors.
\end{itemize}

The evaluation is carried out using the gem5 simulator~\cite{gem5-orig, gem5-2020}, extended with an accurate model of a decoupled vector processor integrated with a superscalar out-of-order core, both conforming to the RISC-V ISA. The benchmark suite includes various attention layer configurations, covering a range of sequence lengths and head dimensions to reflect diverse workload patterns commonly found in transformer models. The results demonstrate three key outcomes: first, the proposed vectorized FlashAttention implementation delivers significant performance gains, confirming its suitability for vector-based architectures; second, the choice of an effective tiling strategy proves crucial for maximizing memory reuse and vector unit efficiency; and third, increasing the available vector length consistently improves performance by exposing greater data-level parallelism, highlighting the scalability of the approach on modern vector hardware.

\section{Baseline Attention and FlashAttention algorithm}

The attention mechanism allows models to assign varying levels of importance to elements within an input sequence, making it possible to capture long-range dependencies and contextual associations. It operates through the interaction of query, key, and value vectors, all of which are generated from the input embeddings~\cite{base_attn}.
In practice, the attention mechanism operates across multiple heads in parallel, known as multi-head attention, allowing the model to capture complex relationships more effectively~\cite{base_attn}.  
In the following, without loss of generality, we will limit our discussion to a single-head attention.

\subsection{Baseline Attention with Lazy Softmax Division}

Given a query vector $\vec{q}$ and sets of key and value vectors defined as $K = \vec{k}_1, \ldots, \vec{k}_N$ and $V = \vec{v}_1, \ldots, \vec{v}_N$, the attention mechanism is computed as follows:
\begin{equation}
s_i = \text{dot}(\vec{q}, \vec{k}_i)\qquad f_i = \frac{e^{s_i}}{\sum_j e^{s_j}}\qquad \text{Attn}(\vec{q}, \text{K}, \text{V}) = \sum_i f_i, \vec{v}_i
\nonumber
\end{equation}

Here, each attention score $s_i$ measures the similarity between the query vector and the $i$th key vector through a dot product. These scores are then passed through a softmax function to determine the relevance of each token to the query. This step involves computing the exponential of each score and normalizing by the sum of all exponentials, resulting in a probability distribution over the input tokens. The final attention output is obtained as a weighted sum of the value vectors, where each weight reflects the relative importance of the corresponding token.

However, computing exponentials of large scores can cause numerical instability, potentially leading to overflow. To address this, a numerically stable variant known as the safe softmax is used. This method subtracts the maximum score from all individual scores before exponentiation, preserving the behavior of softmax while preventing overflow:
\begin{equation}
f_i = \frac{e^{s_i - \max}}{\sum_j e^{s_j - \max}} \quad \text{(safe softmax)}
\nonumber
\end{equation}

Instead of performing exponentiation, normalization, and weighted summation as separate steps, lazy softmax architectures~\cite{lazy_softmax, elsa} take a different approach. They accumulate both the weighted sum of the value vectors and the sum of exponentials in parallel, delaying the division step until the end:
\begin{equation}
s_i = \text{dot}(\vec{q}, \vec{k}_i)\quad f_i = e^{s_i - \max}\quad
\text{Attn}(\vec{q}, \text{K}, \text{V}) = \frac{\sum_i f_i \vec{v}_i}{\sum_j e^{s_j - \max}}
\nonumber
\end{equation}

Alg.~\ref{alg:attn-lazy-scalar} outlines the computation of attention using the lazy softmax approach. The process begins by calculating the dot product between the query and each key vector, while simultaneously tracking the maximum score for numerical stability. Then, the output vector $\vec{o}_i$ is built incrementally by summing the products of each value vector and its corresponding exponentiated, shifted score. At the same time, the running total of exponentials, denoted as $\ell_i$, is updated. The final attention result is produced by dividing the accumulated weighted sum by the total sum of exponentials $\ell_N$.

\begin{algorithm}[t]
\caption{Attention with lazy softmax division}
\label{alg:attn-lazy-scalar}
\begin{algorithmic}[1]
\ForEach {query $\vec{q}$}
\For{$i = 1:N$} 
\State $s_i \gets \text{dot}(\vec{q}, \vec{k}_i)$
\State $m_i \gets \max(m_{i-1}, s_i)$
\EndFor
\State $\ell_0 \gets 0$
\For{$i = 1:N$} 
\State $\vec{o}_i \gets \vec{o}_{i-1} + e^{s_i - m_N}\cdot \vec{v}_i$
\State $\ell_i \gets \ell_{i-1} +e^{s_i - m_N}$
\EndFor
\State $\text{attn}(\vec{q}, K, V) \gets \vec{o}_N/\ell_N$
\EndFor
\end{algorithmic}
\end{algorithm}

The attention computation in Alg.~\ref{alg:attn-lazy-scalar} has a major bottleneck, as output calculation can only start after determining the maximum attention score, limiting efficiency for large sequence lengths 
($N$). FlashAttention~\cite{fa}, inspired by online softmax computation~\cite{online-softmax}, addresses this by merging the two inner loops of Alg.~\ref{alg:attn-lazy-scalar} into a single loop that computes all necessary variables online.

\begin{algorithm}[t]
\caption{FlashAttention with delayed softmax division}\label{alg:flash-attn2}
\begin{algorithmic}[1]
\ForEach {query $\vec{q}$}
\For{$i = 1:N$} 
\State $s_i \gets \text{dot}(\vec{q}, \vec{k}_i)$
\State $m_i \gets \max(m_{i-1}, s_i)$
\State $\ell_i \gets \ell_{i-1}e^{m_{i-1}-m_i}+e^{s_i-m_i}$
\State $\vec{o}_i \gets \vec{o}_{i-1} e^{m_{i-1}-m_i}+e^{s_i-m_i}\vec{v}_i $
\EndFor
\State $\text{attn}(\vec{q}, K, V) \gets \vec{o}_N/\ell_N$
\EndFor
\end{algorithmic}
\end{algorithm}

\subsection{FlashAttention Algorithm}
FlashAttention~\cite{fa} transforms attention computation into an online process. The main difference between FlashAttention and the baseline attention mechanism (Alg.~\ref{alg:attn-lazy-scalar}) is that FlashAttention performs all required computations within a single inner loop. This eliminates the need to precompute the maximum attention score and avoids storing alignment scores in memory~\cite{blockwise}. In effect, it addresses the scalability limitations of Alg.\ref{alg:attn-lazy-scalar} as sequence length increases. 

The FlashAttention kernel is presented in Alg.~\ref{alg:flash-attn2} following a vector-oriented description of the corresponding algorithm. 
Specifically, Alg.~\ref{alg:flash-attn2} refers to the improved version of FlashAttention~\cite{fa2}, \emph{aka}, FlashAttention-2. The next version, i.e., FlashAttention-3~\cite{fa-3} follows exactly the same algorithmic structure and interleaves block‑wise GEMM and softmax computation that is specific for GPU implementations.
In each iteration of FlashAttention, the dot product between the query vector and the key vectors produces a similarity score $s_i$. The maximum attention score up to the current step $m_i$, is also identified. 
The multiplication by \( e^{m_{i-1} - m_i} \) in the computation of \( \ell_i \) in line 5 of Alg.~\ref{alg:flash-attn2}, adjusts the previously accumulated sum of exponentiated scores to the new maximum attention score. 
In line 6, The output vector \( o_i \) is updated by adding the new value vector \( V \), weighted by its softmax importance, to the previous output vector \( o_{i-1} \), which is similarly adjusted by the scaling factor \( e^{m_{i-1} - m_i} \). Finally, in line 8, the attention of a query vector is computed by dividing the output \( o_N \) by the sum of all exponentiated attention scores $\ell_N$.

As shown in Alg.~\ref{alg:flash-attn2}, parallelization can be applied either to the outer or the inner loop. Parallelizing the outer loop allows for processing multiple queries for the same column of \( K \). However, in this case, all state needed per query vector such the maximum attention score $m_i$ and the running sum of exponents $\ell_i$ should be also stored. Parallelizing the inner loop matches better to the vectorization of FlashAttention that is the focus of this work.

\begin{algorithm}[t]
\caption{FlashAttention applied at blocks of key matrix}
\label{alg:flash-attn2-blocks}
\begin{algorithmic}[1]
\ForEach {query $\vec{q}$}
\For{$i = b:b:N$} 
\State $\vec{s} \gets \vec{q} \cdot K^T[:,\ (i-b):i-1]$
\State $m_i \gets \max(m_{i-b}, \vec{s_i})$
\State $\ell_i \gets \ell_{i-b}e^{m_{i-b}-m_i}+\sum_{j=0}^{b-1} e^{s_{i-j}-m_i}$
\State $\vec{o}_i \gets \vec{o}_{i-b} e^{m_{i-b}-m_i}+\sum_{j=0}^{b-1} e^{s_{i-j}-m_i} V[i-j-1,\ :]$%\vec{v}_{(j+i-b)}$
\EndFor
\State $\text{attn}(\vec{q}, K, V) \gets \vec{o}_{N}/\ell_{N}$
\EndFor
\end{algorithmic}
\end{algorithm}

In this case, instead of computing a scalar alignment score \( s_i \), we compute a vector of alignment scores \( \vec{s} \) by performing a dot product across multiple columns of \( K \) and the same query vector $\vec{q}$. In this way, the maximum score identification, the computation of sum of exponents, and output accumulation should also adapt to operate on this vector of attention scores. This adaptation can be easily derived by unrolling the recursive computation of $m_i$, $\ell_i$ and $o_i$. For instance, by unrolling once the running sum of exponents $\ell_i$ we get that:
\begin{align}
\ell_i & = \ell_{i-1} e^{m_{i-1}-m_i} + e^{s_i-m_i} \nonumber\\
       & = \left ( \ell_{i-2} e^{m_{i-2}-m_{i-1}} + e^{s_{i-1}-m_{i-1}} \right )
          e^{m_{i-1}-m_i} + e^{s_i-m_i} \nonumber\\
       & =  \ell_{i-2} e^{m_{i-2} - m_i} + e^{s_{i-1}-m_i} +  e^{s_i-m_i} \nonumber
\end{align}
Repeating the same procedure $b$ times leads to the following lookahead form for computing the sum of exponentials for $b$ consecutive attention scores.
\begin{align}
 \ell_i      & =  \ell_{i-b} e^{m_{i-b} - m_i} + e^{s_{i-b+1}-m_i} + \ldots +  e^{s_i-m_i} \nonumber\\
       & = \ell_{i-b} e^{m_{i-b} - m_i} + \sum_{j=0}^{b-1} e^{s_{i-j}-m_i} 
\label{e:unroll}
\end{align}

From~\eqref{e:unroll}, it can be easily seen that $\ell_i$ can be computed by the sum of exponentials of $b$ attention scores $s_i, s_{i-1}, \ldots s_{i-b}$ reduced by the current maximum score $m_i$. Their sum is then accumulated to the adjusted $\ell_{i-b}$ sum. This adjustment depends only on the current maximum $m_i$ and the previous block maximum $m_{i-b}$. Likewise, the output accumulation $\vec{o}_i$ can be performed at the vector level including the multiplication with the value vectors and without requiring extra intermediate adjustments to the maximum value. This vector-level lookahead mechanism in the FlashAttention algorithm is illustrated in Alg.~\ref{alg:flash-attn2-blocks}.

The application of blocking for the query, key, and value matrices, introduced in the original FlashAttention proposals~\cite{fa, fa2}, will be examined in Section~\ref{s:tiling} in the context of the vectorized implementation of FlashAttention.

\section{Vectorization of FlashAttention}
The proposed vectorization of FlashAttention using the RISCV ISA follows the block-level execution described in Alg.~\ref{alg:flash-attn2-blocks}.

The execution of FlashAttention of one query vector is split in five parts: In the first part, the vector of attention scores $\vec{s}$ is computed using a vector$\times$matrix row-based vectorized multiplication. The vector corresponds to the query vector and the matrix is formed by $b$ consecutive columns of the Key matrix. In the second part, the maximum attention in $\vec{s}$ is identified in parallel and the corresponding exponentials are computed in a fully-vectorized form using $\vec{s}$ and the current and the previous maximum attention scores. The third part refers to the computation of the sum of exponents $\ell_i$ the output vector  $\vec{o}_i$. The fourth part corrects the maximum values for the previous sum of exponent $\ell_{i-b}$ and the output vector $o_{i-b}$ and the last part computes for the current query vector the attention vector by dividing the accumulated output with the sum of exponents of all attention scores.

The execution of each part using vectorized operations that correspond closely to typical RISCV Vector ISA is shown in Alg.~\ref{alg:fa}. In the presented algorithm we assume that the vector processor operates on an arbitrary large vector length that matches the head dimension of the attention model, denoted as $d$. The more realistic scenario, where vector length is smaller than the head dimension is discussed in Section~\ref{s:tiling}. 

\begin{algorithm}[t!]
\caption{Vectorized FlashAttention for RISCV Vector ISA}
\label{alg:fa}
\begin{algorithmic}[1]
\ForEach {query $\vec{q}$}
\State vload  $\text{v}_q$, $\vec{q}$
\For{$i = b:b:N$} 
\State $\text{v}_s \gets 0$
\For{$j=0: d-1$}
    \State vload $\text{v}_k$, $K^T[j,\ (i-b):\ i-1]$
    \State vrgather $\text{v}_{a}, \ \text{v}_{q}, \ j$
    \algorithmiccomment{Broadcast jth element of $\vec{q}$}
    \State{vmacc $\text{v}_s,\ \text{v}_k,\ \text{v}_{a}$} \algorithmiccomment{Multiply-Accumulate}
\EndFor
\State vredmax $\text{v}_{max},\ \text{v}_s, \ \text{v}_{oldmax}$ \algorithmiccomment{Obtain current maximum value}
\State vrgather $\text{v}_{max}, \ \text{v}_{max}, \ 0$
\State {$\text{v}_b$ = vexp($\text{v}_{oldmax}$ - $\text{v}_{max}$)} \algorithmiccomment{Calculate $e^{m_{i-b}-m_i}$}
\State $\text{v}_c$ = vexp($\text{v}_s$ - $\text{v}_{max}$) 
\algorithmiccomment{Calculate $e^{\vec{s}-m_i}$}
\State vredsum $\text{v}_{sum}, \ \text{v}_c$  
\algorithmiccomment{Calculate current sum of exponents}
\State vrgather $\text{v}_{sum}, \ \text{v}_{sum}, \ 0$
\algorithmiccomment{Broadcast sum of exponents}
\State {vmsneq $\text{v}_{mask}, \ \text{v}_{oldmax}, \ \text{v}_{max}$}\algorithmiccomment{Create mask to perform lookahead}
\For{$j=0 : b-1$}
\State vload $\text{v}_v$,  $V[i-j-1,\ :]$ \algorithmiccomment{Load row of V}
\State vrgather $\text{v}_{d}, \ \text{v}_{c}, \ (b-1-j)$
\State{vmacc $\text{v}_{out},\ \text{v}_v,\ \text{v}_{d}$}
\EndFor
\State {vmul $\text{v}_{oldsum},\ \text{v}_{oldsum},\ \text{v}_b,\ \text{v}_{mask}$} \algorithmiccomment{Masked multiplication}
\State vadd $\text{v}_{sum}, \ \text{v}_{oldsum}, \ \text{v}_{sum}$ 
\algorithmiccomment{Calculate $l_i$}
\State vmv $\text{v}_{oldsum}, \ \text{v}_{sum}$
\algorithmiccomment{Move $l_{i-b}$ to $l_{i}$}
\State {vmul $\text{v}_{oldout},\ \text{v}_{oldout},\ \text{v}_b,\ \text{v}_{mask}$} 
\algorithmiccomment{Masked multiplication}
\State vadd $\text{v}_{out}, \ \text{v}_{oldout}, \ \text{v}_{out}$ 
\algorithmiccomment{Calculate $o_i$}
\State vmv $\text{v}_{oldmax}, \ \text{v}_{max}$
\algorithmiccomment{Move $m_{i-b}$ to $m_{i}$}
\EndFor
\State {vdiv $\text{v}_{out},\ \text{v}_{out}, \ \text{v}_{sum}$}
\algorithmiccomment{Perform $\vec{o_N}/\ell_N$}
\State vstore $\text{v}_{out}, \vec{o}_N$
\algorithmiccomment{Store $\vec{o}_N$}
\EndFor
\end{algorithmic}
\end{algorithm}

To compute the vector of alignment scores, we utilize the row-wise matrix multiplication~\cite{matraptor}, which is highly efficient for vector-matrix multiplication. This operation requires each element of the vector $\text{v}_q$ to be multiplied with the corresponding row vector $\text{v}_k$ from the transposed key matrix $K^T$. To accomplish this, during each iteration of the loop in line 5 of Alg.~\ref{alg:fa}, we must isolate individual elements of $\text{v}_q$. For this purpose, we employ the \texttt{vrgather} instruction, which enables indexing into $\text{v}_q$ and broadcasting a selected element across all elements of the vector $\text{v}_a$. The alignment score vector is then computed using a multiply-and-accumulate instruction, as shown in line 8 of Alg.~\ref{alg:fa}, completing the first phase of FlashAttention’s execution for a single query vector.

It is worth noting that, instead of the row-wise approach, a dot-product-based method could have been used. However, such an approach would generate multiple scalar values that would need to be rearranged into a single vector. This process would require several additional shift and merge operations, resulting in a greater number of inefficient vector instructions. In contrast, the row-wise multiplication algorithm directly places each computed value in its correct position within the result vector, thus minimizing the number of vector instructions required and improving overall efficiency.

In the second part, covering lines 10–13, the maximum value is computed across all elements of the vector $\text{v}s$ and the previously calculated maximum that is stored in vector $\text{v}{oldmax}$. This is achieved using the \texttt{vredmax} instruction, which, in the RISC-V ISA, stores the resulting maximum in the first element of the vector $\text{v}{max}$. To propagate this value across the entire vector, it is broadcast to all elements of $\text{v}{max}$ using the \texttt{vrgather} instruction, as shown in line 11 of Alg.~\ref{alg:fa}.

Moreover, since the RISC-V ISA lacks direct support for vector exponentiation instructions, the operations in lines 12 and 13 of Alg.~\ref{alg:fa}, which compute the terms $e^{m_{i-b} - m_i}$ and $e^{\vec{s} - m_i}$, respectively, are conceptually referred to as vexp for clarity. The implementation details of these operations will be discussed in the following subsection.

In the third part, the sum of exponents is computed using the \texttt{vredsum} instruction in line 14 of Alg.~\ref{alg:fa}. Then, in lines 17–21, the current output vector $\vec{o}_i$ is computed and stored in the vector $\text{v}_{out}$ by applying the row-wise multiplication algorithm between the vector of exponentiated attention scores and the corresponding block of the value matrix. By the end of this part, $\text{v}_{sum}$ contains the term $\sum_{j} e^{s_{i-j} - m_i}$, which is required in line 5 of Alg.~\ref{alg:flash-attn2-blocks}, and $\text{v}_{out}$ holds the term $\sum_{j} e^{s_{i-j} - m_i} V[i-j-1,\ :]$, as defined in line 6 of Alg.~\ref{alg:flash-attn2-blocks}.

To complete the computation described in lines 5 and 6 of Alg.~\ref{alg:flash-attn2-blocks}, the previously computed sum of exponents $\ell_{i-b}$ and output vector $o_{i-b}$ must be added to the newly computed terms, after applying a correction factor $e^{m_{i-b}-m_i}$. To simplify this correction, we use a conditional assignment. The correction is necessary whenever the current maximum value $m_i$ differs from the previous maximum $m_{i-b}$. The result of this comparison is stored in line 16 in a vector mask. This mask controls the subsequent masked multiplication instructions that adjust the vectors $\text{v}_{oldsum}$ and $\text{v}_{oldout}$ based on the current maximum value, as shown in lines 22 and 25 of Alg.~\ref{alg:fa}, respectively. The corrected vectors are subsequently added to the current sum of exponents and the output vector $\vec{o}_i$, as illustrated in lines 23 and 26 of Alg.~\ref{alg:fa}.

Finally, in the last part, the output vector $\vec{o}_i$ is normalized by dividing it by the final sum of exponents $\ell_i$, as shown in line 29 of Alg.~\ref{alg:fa}.

\subsection{Exponent approximation}

To complete the FlashAttention kernel we need to approximate in lines 12 and 13 of Alg.~\ref{alg:fa} the exponentials of the two maximum differences attention scores. The resulting exponentiated scores should return in the same floatint-point format as the attention score differences. In RISC-V Vector ISA there is no direct instruction to compute exponentials in vector form. This absence means that software implementations must rely on approximation techniques to perform efficiently this computation at the vector level.

In the case of vectorized FlashAttention, a generic exponential function approximator is unnecessary and a custom approximation suffices, as we only need to compute attention score differences that are zero or negative. This stems from the fact that attention scores are always offset by the maximum score. Therefore, the first step is to appropriately quantize the attention score differences. Then, using the approach proposed by Blin~\cite{blin}, we compute the exponential of the quantized differences in a single step and directly convert the result into floating-point format.

\subsubsection{Quantization of attention score differences}
In general, converting a floating-point number to its fixed-point representation requires many integer bits to accommodate the wide range of floating-point arithmetic~\cite{wide-acc}. However, in this case, we can safely clip the attention score differences to a much smaller dynamic range, as it will only be used for the computation of $e^x$ for negative values of $x$. In this range, $e^x$ quickly converges to 0, even for small negative $x$ values, i.e., $e^{-15}=3.1\cdot 10^{-7}$.

In this work, we employ scaling quantization of the clipped attention score differences. 
Each attention score difference $\Delta$ is scaled by \( s \) and rounded. The scaling factor is defined as:
\[
s = \frac{2^{b - 1} - 1}{a},
\]
where \( b \) is the total number of bits in the fixed-point representation, and \( a \) is the largest absolute value to be represented. 
The quantized fixed-point value is then given by:
\begin{equation}
x_f = \operatorname{int}(s \cdot \Delta).
\label{eq:xf}
\end{equation}

\noindent For simplicity, we set \( b = 32 \) and choose the maximum absolute value \( a \) to be 256. This corresponds to a fixed-point representation with a 9-bit integer part and a 23-bit fractional part.

\subsubsection{Exponential of quantized attention score differences}
To simplify computation, it is preferable to work with powers of two rather than natural exponents. Since $e^x = 2^{x \log_2 e}$, our goal is to compute $2^{x_f \log_2 e}$, where $x_f$ is the quantized attention score difference computed in~\eqref{eq:xf}, and obtain the result directly in floating-point format without any additional conversion step.

Leaving aside for now the multiplication of $x_f$ by $\log_2 e$, we note that since $x_f$ is a fixed-point number with an integer part $I_x$ and a fractional part $F_x$, we can express $2^{x_f}$ as follows:
\begin{equation}
2^{x_f} = 2^{I_x + F_x} = 2^{I_x} \cdot 2^{F_x}
\label{e:pow-2-quant-x}
\end{equation}
Since $F_x$, being a fractional part, lies between 0 and 1 by construction, we can safely approximate $2^{F_x}$ by $1 + F_x$. Substituting this approximation into~\eqref{e:pow-2-quant-x}, we obtain:
\begin{equation}
2^{x_f} \approx 2^{I_x} \cdot (1 + F_x)
\end{equation}
With this approximation, we observe that the result of $2^{x_f}$ takes the form of a floating-point number, where $I_x$ corresponds to the unbiased exponent and $F_x$ to the mantissa.

To fit to the floating-point representation the exponent bias must be restored affecting only the integer part $I_x$ of $x_f$ and leaving the fraction part $F_x$ unchanged. As noted by Blin~\cite{blin}, this can be achieved by adding to $x_f$ bias $B$ scaled by the fraction (mantissa) bit width $W_M$:
\begin{align}
2^{x_f} \approx 2^{I_x-B} \cdot (1 + F_x) = x_f + B \cdot 2^{W_M}
\label{e:formula-1}
\end{align}
This addition effectively incorporates the exponent bias back into \( I_x \), thereby reconstructing the correct binary representation of the floating-point value of \( 2^x_f \). 

\begin{algorithm}[t]
\caption{Vectorized exponentiaion of attention score differences in FlashAttention}
\label{a:vexp}
\begin{algorithmic}[1]
\State vmfle $\text{v}_{mask},\ \text{v}_{exp},\ -15$
\algorithmiccomment{Create mask for clipping}
\State vmul $\text{v}_{exp}, \ s \cdot \log_2(e)$
\algorithmiccomment{Floating point multiplication}
\State $\text{v}_{exp}$ = int($\text{v}_{exp}$)
\algorithmiccomment{Integer conversion}
\State vadd $\text{v}_{exp},\ \text{v}_{exp}, \ B \cdot 2^{W_M}$
\algorithmiccomment{Integer addition}
\State vadd $\text{v}_{exp},\ \text{v}_{zero},\ e^{-15},\text{v}_{mask}$
\algorithmiccomment{Replace the clipped values of x with $e^{-15}$}
\end{algorithmic}
\end{algorithm}

This implementation of vectorized exponentiation is presented in Alg.~\ref{a:vexp}. Let \( \mathbf{v}_{\text{exp}} \) denote the input vector containing the values that need to be exponentiated. The computation begins by identifying elements in \( \mathbf{v}_{\text{exp}} \) that are less than \(-15\). For these elements, a mask is generated to indicate which values require clipping.

Following the mask generation, a floating-point multiplication is performed using the precomputed constant \( s \cdot \log_2(e) \), effectively computing the product \( s \cdot x_{\text{exp}} \). This result is then converted to an integer representation, as shown in line 2 of Alg.~\ref{a:vexp}. An integer addition is subsequently performed between the converted result and the constant \( B \cdot 2^{W_M} \).

Finally, elements of \( \mathbf{v}_{\text{exp}} \) that were marked for clipping are replaced. This is achieved via a masked vector operation that conditionally assigns the value \( e^{-15} \) to the corresponding positions, using a zero-initialized vector \( \mathbf{v}_{\text{zero}} \) to facilitate the update. For the elements not requiring clipping, the addition will not have an effect due to masking.
The final output vector approximates the exponential function \( e^x \) for the attention score differences in the context of FlashAttention.

The five operations described in Alg.~\ref{a:vexp}, i.e., masking operation, floating-point multiplication, integer conversion, and addition, are all supported by standard RISC-V vector instructions and allows us to compute exponential functions as needed in the context of Flashattention in a cost-efficient manner. 

\section{Tiling in Vectorized FlashAttention}
\label{s:tiling}

Since a vector engine is equipped with a finite number of arithmetic logic units (ALUs) and can operate on only a limited amount of data concurrently, the size of the input arrays often exceeds the level of parallelism the hardware can natively support. This mismatch between data volume and computational capacity necessitates the use of additional techniques to manage and optimize performance. One such method is tiling, which partitions the data into smaller, manageable blocks that fit within the engine's parallel processing capabilities. By processing these tiles sequentially or in batches, the system ensures that the full attention output is computed correctly and efficiently.

\begin{algorithm}[!ht]
\caption{Vectorized FlashAttention-2 with d greater than vector length}
\label{alg:fa-d}
\begin{algorithmic}[1]
\ForEach {query $\vec{q}$}
\For{$i = vl:vl:N$} 
\State $\text{v}_s \gets 0$
\For{$h=0:d/vl-1$}
\State \textcolor{blue}{vload  $\text{v}_q$, $\vec{q}\ [h\cdot vl: \ (h+1)\cdot vl-1]$}
\For{$j=0: vl-1$}
    \State \textcolor{blue}{vload $\text{v}_k$, $K^T[h\cdot vl+j,\ (i-vl):\ i-1]$}
    \State vrgather $\text{v}_{a}, \ \text{v}_{q}, \ j$
    \algorithmiccomment{Broadcast jth element of $\vec{q}$}
    \State{vmacc $\text{v}_s,\ \text{v}_k,\ \text{v}_{a}$} \algorithmiccomment{Multiply-Accumulate}
\EndFor
\EndFor
\State vredmax $\text{v}_{max},\ \text{v}_s, \ \text{v}_{oldmax}$ \algorithmiccomment{Obtain current maximum value}
\State vrgather $\text{v}_{max}, \ \text{v}_{max}, \ 0$
\State {$\text{v}_b$ = vexp($\text{v}_{oldmax}$ - $\text{v}_{max}$)} \algorithmiccomment{Calculate $e^{m_{i-b}-m_i}$}
\State $\text{v}_c$ = vexp($\text{v}_s$ - $\text{v}_{max}$) 
\algorithmiccomment{Calculate $e^{\vec{s}-m_i}$}
\State vredsum $\text{v}_{sum}, \ \text{v}_c$  
\algorithmiccomment{Calculate current sum of exponents}
\State vrgather $\text{v}_{sum}, \ \text{v}_{sum}, \ 0$
\algorithmiccomment{Broadcast sum of exponents}
\State {vmsneq $\text{v}_{mask}, \ \text{v}_{oldmax}, \ \text{v}_{max}$}\algorithmiccomment{Create mask to perform lookahead}
\For{$h=0:d/vl-1$}
\For{$j=0 : vl-1$}
\State \textcolor{blue}{vload $\text{v}_v$,  $V[i-j-1,\ h\cdot vl: \ (h+1)\cdot vl-1]$} \algorithmiccomment{Load row of V}
\State vrgather $\text{v}_{d}, \ \text{v}_{c}, \ (vl-1-j)$
\State{vmacc $\text{v}_{out},\ \text{v}_v,\ \text{v}_{d}$}
\EndFor
\State \textcolor{blue}{vload  $\text{v}_{oldout}$, $\vec{o}_N\ [h\cdot vl: \ (h+1)\cdot vl-1]$}
\State {vmul $\text{v}_{oldout},\ \text{v}_{oldout},\ \text{v}_b,\ \text{v}_{mask}$} 
\algorithmiccomment{Masked multiplication}
\State vadd $\text{v}_{out}, \ \text{v}_{oldout}, \ \text{v}_{out}$ 
\algorithmiccomment{Calculate $o_i$}
\State \textcolor{blue}{vstore  $\text{v}_{out}$, $\vec{o}_N\ [h\cdot vl: \ (h+1)\cdot vl-1]$}
\EndFor
\State {vmul $\text{v}_{oldsum},\ \text{v}_{oldsum},\ \text{v}_b,\ \text{v}_{mask}$} \algorithmiccomment{Masked multiplication}
\State vadd $\text{v}_{sum}, \ \text{v}_{oldsum}, \ \text{v}_{sum}$ 
\algorithmiccomment{Calculate $l_i$}
\State vmv $\text{v}_{oldsum}, \ \text{v}_{sum}$
\algorithmiccomment{Move $l_{i-b}$ to $l_{i}$}
\State vmv $\text{v}_{oldmax}, \ \text{v}_{max}$
\algorithmiccomment{Move $m_{i-b}$ to $m_{i}$}
\EndFor
\For{$h=0:d/vl-1$}
\State \textcolor{blue}{vload  $\text{v}_{out}$, $\vec{o}_N\ [h\cdot vl: \ (h+1)\cdot vl-1]$}
\State {vdiv $\text{v}_{out},\ \text{v}_{out}, \ \text{v}_{sum}$}
\algorithmiccomment{Perform $\vec{o_N}/\ell_N$}
\State \textcolor{blue}{vstore $\text{v}_{out}$, $\vec{o}_N\ [h\cdot vl: \ (h+1)\cdot vl-1]$}
\algorithmiccomment{Store $\vec{o}_N$}
\EndFor
\EndFor
\end{algorithmic}
\end{algorithm}

\subsection{Handling arbitrary head dimensions}
The head dimension in attention refers to the size of the query, key, and value vectors used within each attention head. It determines how much information each head can capture independently, and the full model dimension is typically split across multiple heads to allow parallel attention to different representation subspaces.

The head dimension, denoted by \( d \) does not necessarily match the vector length supported by the underlying vector registers. In many cases, \( d \) exceeds the available hardware parallelism, which necessitates the application of tiling techniques to maintain computational efficiency. As shown in Alg.~\ref{alg:fa}, the two intermediate loops depend on the parameters \( d \) and \( b \). While the parameter \( d \) is fixed by the attention mechanism and cannot be altered, the blocking parameter \( b \) is configurable and can be set arbitrarily. In this work, \( b \) is set equal to the vector length (\textit{vl}), thereby aligning the blocking strategy with the parallel processing capabilities of the target vector architecture. 

When head dimension $d$ is larger that the maximum supported vector length we need to introduce tiling in the vectorized FlashAttention. The tiled form of vectorized FlashAttention is presented in Alg.~\ref{alg:fa-d}, with the modifications relative to baseline vectorized FlashAttention (Alg.~\ref{alg:fa}) highlighted in blue. 
Since head dimension exceed maximum vector length, the vectors loaded during computation do not encompass all elements within a full row of the \( Q \), \( K \), and \( V \) matrices. Instead, only a portion of each row is loaded in a given iteration, and the corresponding partial computation is performed. Subsequent iterations then load incrementally the remaining segments of each row, as shown in lines 5, 7 and 21 of Alg.~\ref{alg:fa-d}. Consequently, all intermediate results must be temporarily stored and subsequently reloaded to ensure correctness. Specifically, intermediate load and store operations involving the output vector occur at lines 25 and 28 of Alg.~\ref{alg:fa-d}, as well as at lines 36 and 38, in order to correctly compute the division of the output vector \( \vec{o}_N \) by the sum of exponents \( \ell_N \). It is important to note that if the dimension \( d \) is only slightly larger than the vector length, the loops associated with the head dimension can be unrolled by utilizing additional vector registers. This approach reduces or eliminates the need to store partial results, thereby avoiding these specific loops.

\subsection{Operating in parallel on multiple queries}

Algorithm~\ref{alg:fa-d} adopts a dataflow strategy that processes the entire $K$ and $V$ matrices for each query of matrix $Q$. Although this method yields one row of the attention output at a time, an alternative approach is possible. By processing multiple rows from matrix Q together with corresponding sub-blocks of the $K$ and $V$ matrices in parallel, the residency time of $K$ and $V$ in the cache is extended, achieving better temporal locality. This results in better performance as it will be demonstrated in  Section~\ref{s:exp_res}.

\begin{figure}[t!]
    \centering
    \includegraphics[width=0.85\columnwidth]{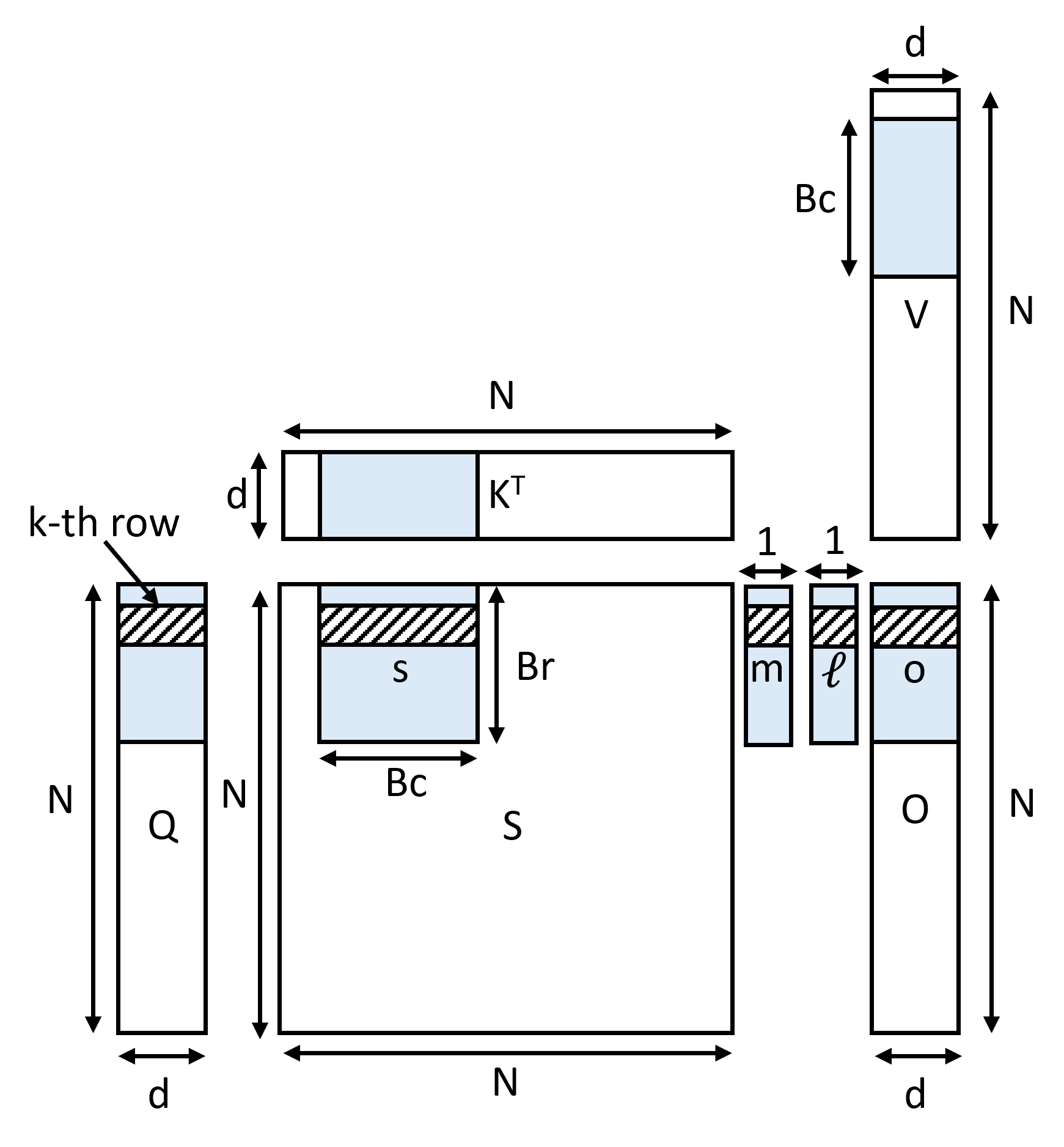}
    \caption{The diagram above illustrates how Attention is computed in blocks. The blue-colored blocks represent the tiled implementation of Attention mechanism. The black-colored blocks correspond to the k-th row of the attention output. N denotes the sequence length, d denotes the head dimension and Br and Bc are the block sizes that can be controlled.}
    \label{f:block fa}
\end{figure}

The strategy of processing multiple rows of matrix \( Q \) is illustrated in Fig.~\ref{f:block fa}, where matrix \( S \) includes the attention scores computed gradually across all query and key vectors. As shown, blocking is applied to the \( Q \), \( K \), and \( V \) matrices. The block size \( B_r \) corresponds to the number of consecutive rows of \( Q \) processed for each block of the \( K \) and \( V \) matrices, while \( B_c \) denotes the number of columns of the \( K \) matrix processed in parallel. This partitioning enables the division of the \( Q \) matrix into \( N / B_r \) blocks, and the \( K \) and \( V \) matrices into \( N / B_c \) blocks. The maximum values and the sums of the exponentiated elements computed for each row are stored, respectively, in the vectors \( m \) and \( \ell \) within each block of size \( B_r \).

\begin{table}[t!]
\centering
\caption{Simulated Processor Configuration}
\label{t:proc-det}
\renewcommand\tabularxcolumn[1]{m{#1}} 
\begin{tabularx}{\columnwidth}{cX}
\hline
Scalar core & 
\begin{itemize}[leftmargin=5pt]
\item 
RISC-V ISA (RV64GC), 8-way-issue out-of-order,
16-entry LSQ, 90 physical integer and 90 physical
 floating-point registers, 60-entry ROB
\item
L1I cache: 1-cycle hit latency, 4-way, 64KB
\item 
L1D cache: 2-cycle hit latency, 4-way, 64KB
\item 
L1I and L1D have 64B cache line
\end{itemize}\\
\hline
Vector engine & \begin{itemize}[leftmargin=5pt]
\item 1024-bit vector engine with 32-lane configuration (32-bit elements $\times$ 32 execution lanes)
\item The vector engine is connected directly to the L2 cache through 16 store queues and 16 load queues
\end{itemize} \\
\hline
L2 cache & \begin{itemize}[leftmargin=5pt]
\item 8-way, 8-bank
\item 8-cycle hit latency, 512KB shared by both the big core and the vector engine
\item 64B cache line
\end{itemize} \\
\hline
Main Memory & DDR4-2400, 19.2 GB/s memory bandwidth\\ 
\hline
\end{tabularx}
\end{table}

\section{Experimental results}
\label{s:exp_res}
The experimental results aim to explore the design space of vectorized FlashAttention and identify the set of parameters that yield the best performance. 

All experiments are conducted using a fully implemented decoupled vector unit integrated with an out-of-order superscalar processor, corresponding to model 1bDV in~\cite{cornell-vector}. The use of a high-performance superscalar core alongside the vector engine is intended to ensure that the scalar portion of the processor does not become a limiting factor in evaluating vector engine performance. Essentially, this configuration allows the vector unit's performance to be measured in isolation, free from interference or bottlenecks introduced by the scalar side. The system is modeled using the gem5 simulator~\cite{gem5-orig}, and the key architectural parameters of the simulated processor are listed in Table~\ref{t:proc-det}

Algorithm~\ref{alg:fa-d} was executed on attention layers of state-of-the-art LLM architectures available on HuggingFace~\cite{hf}. Table~\ref{t:llm} reports the size of the hidden dimension, number of attention heads, and per-head dimensionality for the first attention layer of each model. In transformer architectures, the hidden size refers to the dimensionality of token embeddings and internal hidden states—that is, the number of features used to represent each token at any layer. A larger hidden size enables the model to learn and represent more complex and nuanced patterns. Conversely, the number of attention heads defines how many parallel attention mechanisms are applied to each token. Each head processes information in a distinct subspace of size \texttt{hidden size}\texttt{/}\texttt{num heads}, known as the head dimension, which serves as the primary focus of our evaluations. Notably, the head dimension remains consistent across all layers within each LLM.

\begin{table*}[t]
    \centering
    \caption{The characteristics of contemporary LLM models available on HuggingFace~\cite{hf}}
    \label{t:llm}
    \begin{tabular}{|c|c|c|c|}
    \hline
    \multirow{2}{*}{LLM} & \multicolumn{3}{c|}{Characteristics}\\
    % \cmidrule(l){2-4}
    &Hidden Size & Num. Heads & Head dimension\\
    \hline
    DeepSeek / Qwen-1.5B-L1& 1536& 12& 128\\
    Google / Gemma2-2B-L1 & 2048& 8 & 256\\
    \hline
    \end{tabular}
\end{table*}

\subsection{Speedup achieved through vectorization}

\begin{figure}[h!]
    \centering
    \includegraphics[width=0.8\columnwidth]{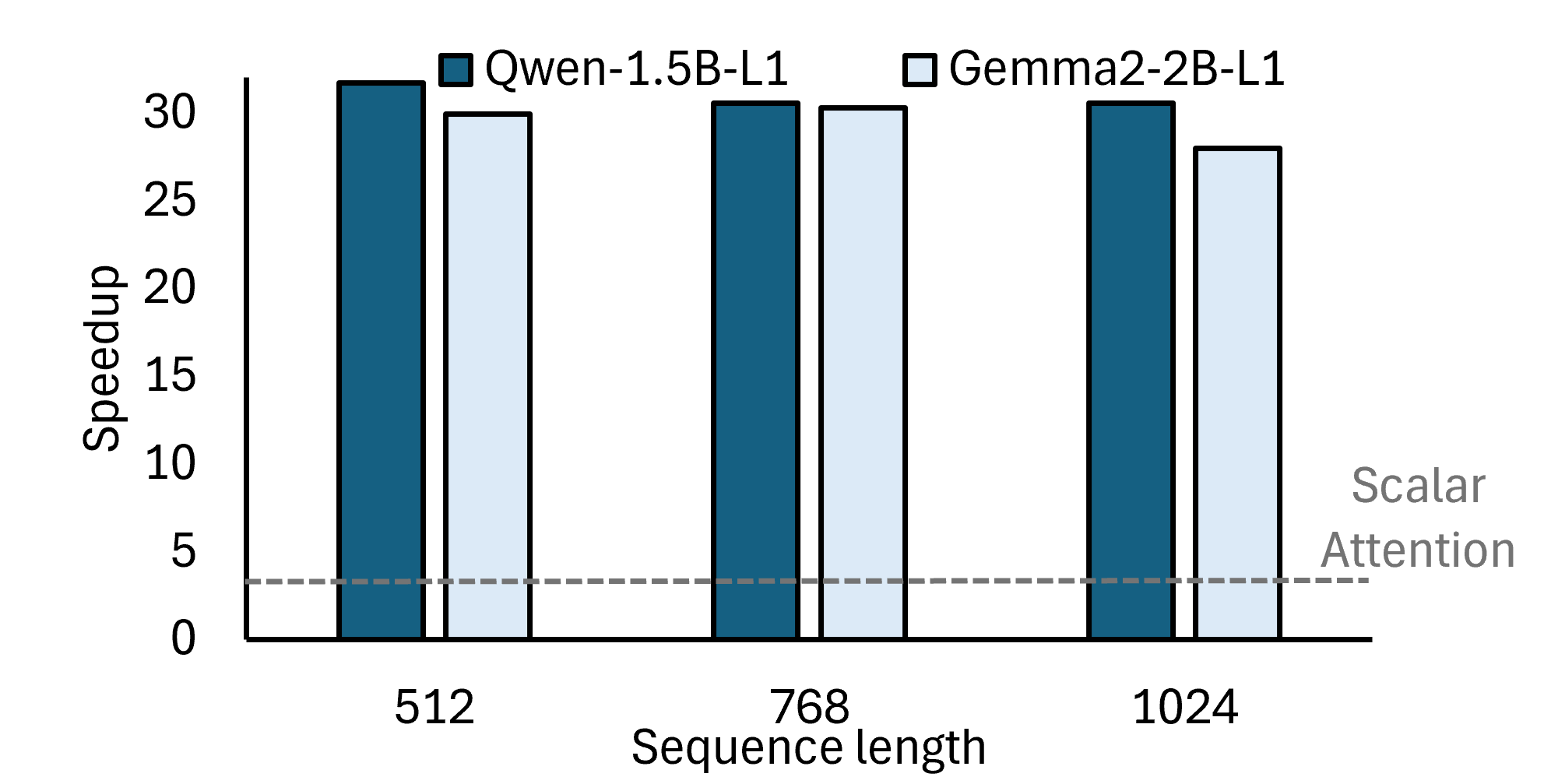}
    \caption{The speedup achieved through the vectorization of Alg.~\ref{alg:fa-d} for \( B_r = 1 \) and vector length 32, compared to each scalar implementation with the same exact configuration for sequence length and head dimension.}
    \label{f:scalar_vs_vector}
\end{figure}

Fig.~\ref{f:scalar_vs_vector} illustrates the performance improvements achieved through code vectorization for a fixed combination of head dimension and sequence length. To evaluate the impact of vectorization, we analyzed a range of sequence lengths with the parameter $B_r$ set to 1 and the parameter $B_c$ set equal to vector length. A vector length of 32 was used, and the reported metrics correspond to the head dimensions of layers from the Qwen-1.5B and Gemma2-2B LLMs. Additionally, we opt to unroll the loops related to head dimension for a factor of 4 to overcome the unnecessary vector memory operations due to the increased size of head dimension. All results presented in Fig.~\ref{f:scalar_vs_vector} are normalized with respect to their corresponding scalar implementations, each having the same sequence length and head dimension as the respective vectorized implementation. As shown, performance approaches its theoretical maximum—aligned with the hardware-supported vector length of 32. However, as the head dimension and sequence length increase, a slight decline in the amount of speedup is observed. This degradation is attributed to the increment on size of all involved arrays, which leads to a greater number of branches and increased pressure on the memory subsystem. The same effect is noticed with the increment of the head dimension, which degrades slightly the performance. The impact of vector length will be examined subsection~\ref{ss:vl}.

\begin{figure}[h]
    \centering
    \includegraphics[width=0.99\columnwidth]{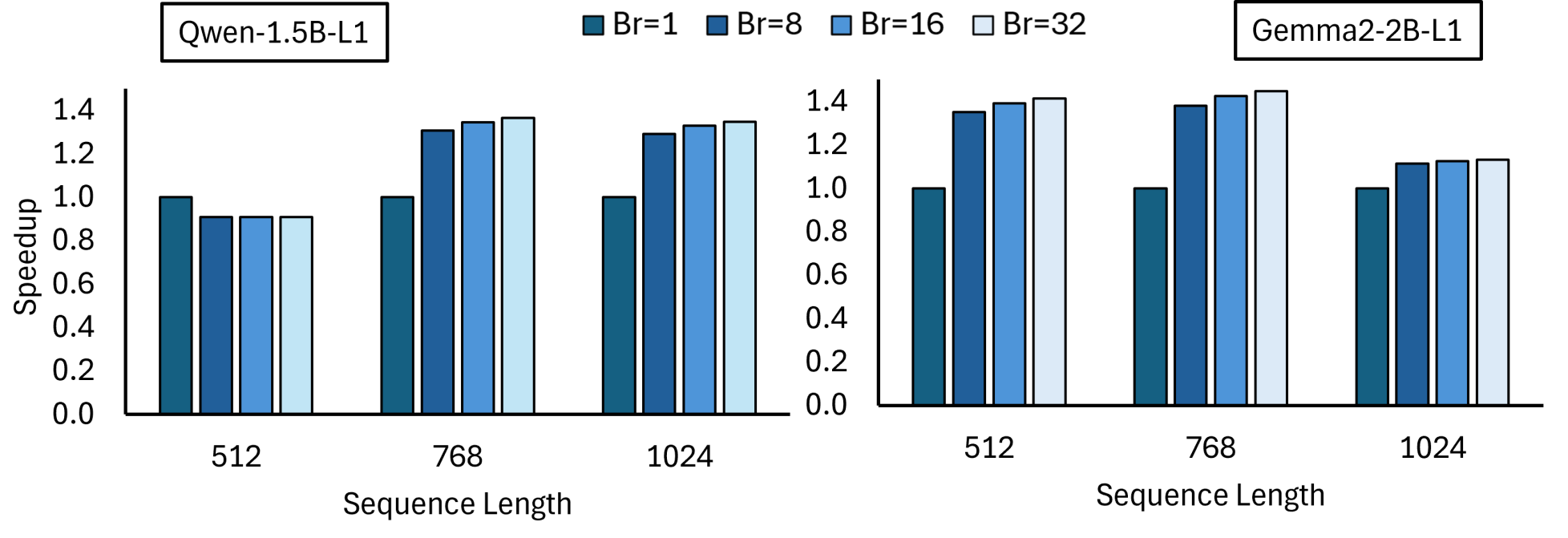}
    \caption{The speedup achieved by the algorithm presented in Alg.~\ref{alg:fa-d} through the use of a $B_r$ parameter greater than 1, compared to the implementation with $B_r = 1$, is evaluated across varying sequence lengths and the head dimensions of the Qwen-1.5B and Gemma2-2B LLMs, with a fixed vector length of 32.}
    \label{f:br_comparison}
\end{figure}

\subsection{The effect of block size in the vectorized performance}

To analyze the impact of the parameter \( B_r \) on the execution cycles of Alg.~\ref{alg:fa-d}, we evaluate its effect under various configurations of sequence length and head dimension, assuming a fixed vector length of 32. All implementations employ loop unrolling by a factor of four in the loops associated with the head dimension.

In Fig.~\ref{f:br_comparison}, the speedup achieved by using a $B_r$ parameter greater than 1 is depicted, with all results normalized relative to the implementation employing a $B_r$ parameter equal to 1. As shown in Fig.~\ref{f:br_comparison}, employing $B_r$ values greater than one generally yields improved performance, often achieving speedup up to 1.4 compared to using $B_r = 1$. 

However, there are notable exceptions, particularly in configurations with a sequence length of 512 and a head dimension of 128, which corresponds to the head dimension for a layer of Qwen-1.5B, as well as those with a sequence length of 1024 and a head dimension of 256, which is the head dimension for a layer of Gemma2-2B. In the latter scenario, the performance improvement is slightly reduced due to increased pressure on the memory subsystem resulting from the larger sequence length and head dimension.

Conversely, in configurations where the sequence length is 512 for the head dimension of the layer of Qwen-1.5B, employing $B_r = 1$ outperforms configurations with larger $B_r$ values. This discrepancy arises because, for $B_r > 1$, the number of memory accesses within inner loops increases. Given the relatively small number of iterations in this setup, the increased memory traffic outweighs the benefits of block reuse, ultimately leading to reduced computational efficiency for the attention mechanism.

These improvements are primarily attributed to the reuse of \( K \) and \( V \) blocks across all rows within a given \( B_r \) block. Specifically, each subsequent block of the \( K \) and \( V \) arrays is accessed only after all rows within the current \( B_r \) block have been processed, thereby enhancing data locality and improving cache utilization. However, as the value of \( B_r \) increases beyond a certain point, the performance benefits begin to diminish due to reduced cache efficiency.

\begin{figure}[t]
    \centering
    \includegraphics[width=0.99\columnwidth]{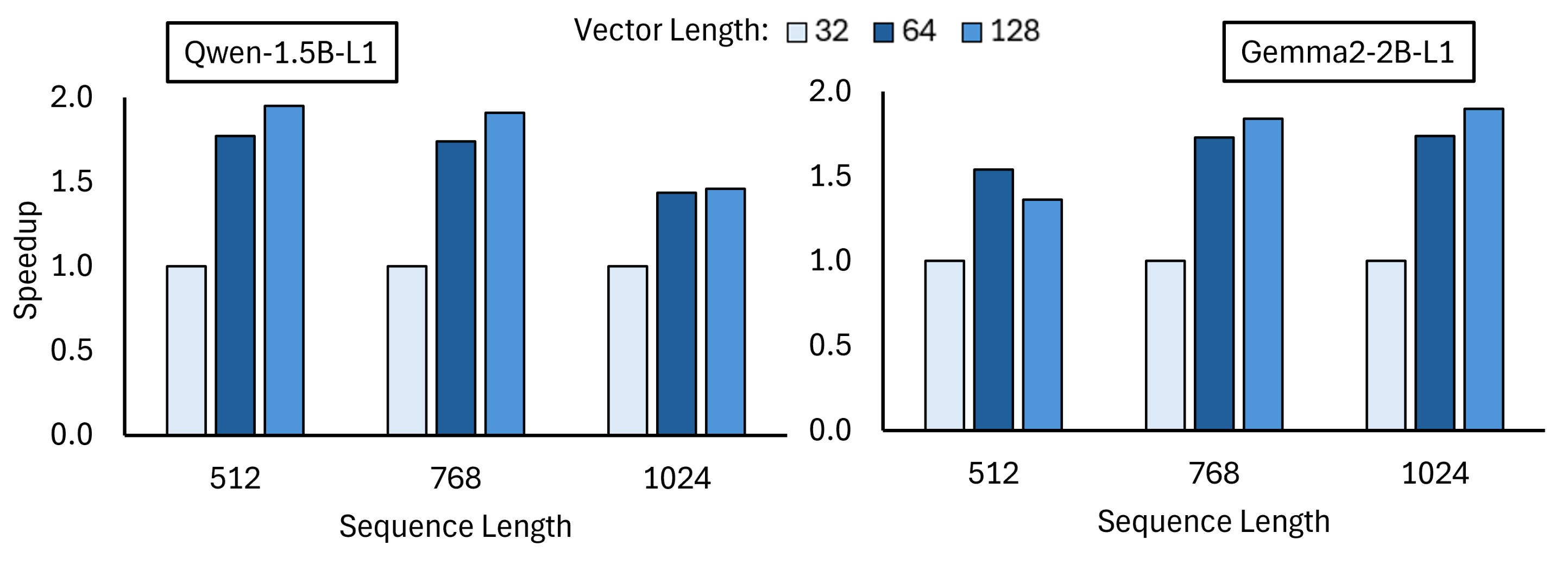}
    \caption{The speedup achieved by Alg.~\ref{alg:fa-d} for three vector length configurations (32, 64, and 128), with the $B_r$ parameter set to 32, is presented. All the results are normalized with respect to the configuration using a vector length of 32.}
    \label{f:vl}
\end{figure}

\subsection{Scaling hardware vector length}
\label{ss:vl}

To assess the impact of hardware vector length, we conducted an analysis of execution cycles across three configurations—32, 64, and 128—while maintaining a constant value of \( B_r = 32 \). This configuration consistently demonstrated optimal overall performance. The maximum loop unrolling related to the head dimension is 4. Figure~\ref{f:vl} depicts the performance gains achieved by increasing vector length relative to the baseline configuration with a vector length of 32. As illustrated, performance improves with higher hardware vector lengths; however, the rate of improvement decreases when the vector length increases. This drop in performance mainly happens because the memory system gets overloaded with more frequent and larger vector load and store operations, which ends up slowing things down overall. Specifically, under conditions where sequence length equals 512 and head dimension equal to the head dimension of the layer of Gemma2-2B, the performance worsens when the vector length is 128 compared to when it is 64. This outcome arises because, with a vector length of 32, covering a head dimension of 256 necessitates 8 vector registers for each loaded vector of Query, Values, and Keys, not accounting for temporary results. This prevents full loop unrolling related to head dimensions and leads to unnecessary loads, which are not beneficial when using a simulated memory subsystem under a vector length of 128

\subsection{Accuracy of exponent approximation on LLM applications}
Having explored the efficiency of the proposed vectorized FlashAttention, in this section we quantify the accuracy loss introduced by the clipping and quantization applied during the exponential approximation of the attention score differences.

We executed ten representative NLP benchmarks from GLUE~\cite{glue} dataset using Google's Flan-T5~\cite{t5} LLM in the FP-32 configuration. For comparison, we considered two implementations: (a) EXP, where the exponential function $e^x$ is computed using the standard math library and (b) EXP-ours where the function of $e^x$ is approximated using the introduced quantization and exponent approximation. To run inference with the LLM we utilized Microsoft's PromptBench workflow~\cite{promptbench}.

The results are shown in Table ~\ref{t:accuracy}. We report the Accuracy and F1-score for all benchmarks with an exception for STS-B, where we report the Pearson Correlation Coefficient (PCC), as it is not a binary classification task. As it can be observed both implementations exhibit comparable performance in terms of inference accuracy and the F1-score metric. In some cases, EXP-ours method even outperforms their full-precision counterparts. While it does not indicate that the model is inherently superior, it confirms that the approximation used to perform vector exponentiation does not degrade performance~\cite{i-bert}.

\begin{table*}[h]
    \centering
    \caption{Performance of Google's FLAN-T5 LLM model on 10 benchmarks from the GLUE dataset~\cite{glue}. Accuracy and F1-score are reported for all benchmarks, except for STS-B, where the Pearson Correlation Coefficient (PCC) is used, as it is not a binary classification task.}
    \begin{tabular}{|l|c|c|c|c|}
        \hline
        \multicolumn{1}{|c|}{\multirow{2}{*}{Benchmarks}} & \multicolumn{2}{c|}{Accuracy (\%)} & \multicolumn{2}{c|}{F1-score} \\
        \multicolumn{1}{|c|}{} & EXP & EXP-ours & EXP & EXP-ours \\
        \hline
        \hline
        \bf STS-2 & 92.1 & 92.1 & 0.921 & 0.931 \\
        \hline
        \bf MNLI-m & 87.5 & 87.5 & 0.794 & 0.800 \\
        \hline
        \bf MNLI-mm & 84.2 & 84.2 & 0.845 & 0.833 \\
        \hline
        \bf QQP & 93.1 & 93.1 & 0.930 & 0.930 \\
        \hline
        \bf QNLI & 93.3 & 93.3 & 0.933 & 0.920 \\
        \hline
        \bf CoLA & 72.0 & 72.0 & 0.830 & 0.840 \\
        \hline
        \bf MRPC & 86.0 & 86.0 & 0.900 & 0.900 \\
        \hline
        \bf RTE & 74.3 & 74.3 & 0.726 & 0.738 \\
        \hline
        \bf WNLI & 62.0 & 62.8 & 0.690 & 0.710 \\
        \hline
        \bf STS-B & 92.0 & 92.0 & - & - \\
        \hline
    \end{tabular}
    \label{t:accuracy}
\end{table*}

\section{Related Work}

The standard attention mechanism, while highly effective, is computationally expensive due to its quadratic complexity with respect to sequence length~\cite{longformer}. This complexity becomes a significant bottleneck in Transformer models, particularly when processing long sequences, as commonly encountered in modern NLP tasks such as document summarization and code generation.

To address these challenges, various strategies have been proposed to improve the efficiency of attention mechanisms. A major line of work focuses on approximating the full attention matrix using sparse~\cite{sparse_attn}, linear~\cite{lin_attn}, or low-rank~\cite{low_rank_attn_2020} attention methods, which aim to strike a balance between computational cost and accuracy. Additionally, techniques like structured sparse pruning~\cite{structured_sparse_attn} have been applied to further reduce complexity by eliminating less significant connections within the attention layers.

Another research direction involves hardware-level acceleration. Custom accelerators~\cite{a3,lazy_softmax} have been developed to optimize core components of attention computation, including matrix operations~\cite{cosa} and softmax evaluation~\cite{softermax}. These hardware efforts aim to enhance speed and scalability, especially for long input sequences. In particular, attention accelerators often store key and value vectors locally in SRAM while streaming query vectors to compute attention scores~\cite{a3,keller,lu}, reducing memory traffic. However, as sequence lengths grow, these vectors must be fetched from slower DRAM, leading to potential performance degradation.

To reduce reliance on memory bandwidth and improve parallelism, in-memory attention computation has also been explored~\cite{x-former}. Moreover, recent designs aim to decouple computational resources from sequence length by accumulating partial softmax results for each column of the attention scores~\cite{lazy_softmax,cosa}. This allows for accurate softmax evaluation without buffering the entire score matrix, thus avoiding memory spills.

The softmax operation itself presents notable performance challenges due to its dependence on reduction operations, such as max-value determination and exponential summation, which limit parallelism. FlashAttention~\cite{fa,fa2,nsquared} addresses these inefficiencies by computing attention in smaller tiles, eliminating the need to store the full attention matrix. Designed for GPUs, FlashAttention recomputes scores on the fly and improves both memory and computational efficiency. FlashAttention-2 further optimizes this by avoiding the need for unified softmax hardware; instead, it performs exponentiation and normalization steps separately. Multiple hardware-based techniques have been explored to implement these non-linear operations efficiently, including piecewise-linear approximations with range reduction~\cite{koca}, logarithmic quantization~\cite{sole}, and other functional approximations~\cite{peano-vit}.

In addition, other optimization techniques aim to reduce latency and energy consumption, such as computation skipping based on token similarity~\cite{elsa,tsacc}, and quantization strategies that minimize data transfer overhead and improve accelerator utilization~\cite{swiftron}.

\section{Conclusions}

The FlashAttention kernel streamlines attention computation by fusing the softmax evaluation. Originally developed for GPUs, this approach enables attention to be computed in tiles of fixed size, independent of the input sequence length. In this work, we aim to adapt the FlashAttention-2 algorithm---an optimized successor to the original FlashAttention---for vector processors, and to demonstrate a fully vectorized implementation. Concurrently, we investigate the impact of varying block configurations over the rows of the \( Q \) matrix on the performance of the attention mechanism. Finally, we examine how different hardware vector lengths influence the efficiency of attention computation under vectorization.

\bibliography{sn-bibliography}% common bib file

\end{document}